\documentclass{article} %
\usepackage{iclr2024_conference,times}

\usepackage{amsmath,amsfonts,bm}

\def\eqref#1{equation~\ref{#1}}

\def\1{\bm{1}}

\DeclareMathAlphabet{\mathsfit}{\encodingdefault}{\sfdefault}{m}{sl}
\SetMathAlphabet{\mathsfit}{bold}{\encodingdefault}{\sfdefault}{bx}{n}

\DeclareMathOperator*{\argmax}{arg\,max}

\usepackage{hyperref}
\usepackage{url}

\usepackage[utf8]{inputenc} %
\usepackage[T1]{fontenc}    %
\usepackage{hyperref}       %
\usepackage{url}            %
\usepackage{booktabs}       %
\usepackage{amsfonts}       %
\usepackage{nicefrac}       %
\usepackage{microtype}      %
\usepackage{xcolor}         %

\usepackage{csquotes}
\MakeOuterQuote{"}
\usepackage{graphicx}
\usepackage{enumitem}
\usepackage{algorithm}
\usepackage[noend]{algpseudocode}
\usepackage{amsmath}
\usepackage{amssymb}
\usepackage{mathtools}
\usepackage{amsthm}
\usepackage{wrapfig}
\usepackage[normalem]{ulem}

\usepackage{subcaption}

\usepackage{listings}
\usepackage{xcolor}

\definecolor{codegreen}{rgb}{0,0.6,0}
\definecolor{codegray}{rgb}{0.5,0.5,0.5}
\definecolor{codepurple}{rgb}{0.58,0,0.82}
\definecolor{backcolour}{rgb}{0.95,0.95,0.92}

\lstdefinestyle{mystyle}{
    backgroundcolor=\color{backcolour},   
    commentstyle=\color{codegreen},
    keywordstyle=\color{magenta},
    numberstyle=\tiny\color{codegray},
    stringstyle=\color{codepurple},
    basicstyle=\ttfamily\footnotesize,
    breakatwhitespace=false,         
    breaklines=true,                 
    captionpos=b,                    
    keepspaces=true,                 
    numbers=left,                    
    numbersep=5pt,                  
    showspaces=false,                
    showstringspaces=false,
    showtabs=false,                  
    tabsize=2
}

\title{DrS: Learning Reusable Dense Rewards\\for Multi-Stage Tasks}

\author{Tongzhou Mu, Minghua Liu, Hao Su \\
UC San Diego\\
\texttt{\{t3mu,mil070,haosu\}@ucsd.edu} 
}

\iclrfinalcopy %
\begin{document}

\maketitle

\begin{abstract}
  The success of many RL techniques heavily relies on human-engineered dense rewards, 
  which typically demands substantial domain expertise and extensive trial and error.
  In our work, we propose \textbf{DrS} (\textbf{D}ense \textbf{r}eward learning from \textbf{S}tages), a novel approach for learning \textit{reusable} dense rewards for multi-stage tasks in a data-driven manner. 
  By leveraging the stage structures of the task, DrS learns a high-quality dense reward from sparse rewards and demonstrations if given. The learned rewards can be \textit{reused} in unseen tasks, thus reducing the human effort for reward engineering. 
  Extensive experiments on three physical robot manipulation task families with 1000+ task variants demonstrate that our learned rewards can be reused in unseen tasks, resulting in improved performance and sample efficiency of RL algorithms. The learned rewards even achieve comparable performance to human-engineered rewards on some tasks. See our \href{https://sites.google.com/view/iclr24drs}{\color{green} project page} for more details.
  \end{abstract}

\section{Introduction}

The success of many reinforcement learning (RL) techniques heavily relies on dense reward functions \citep{hwangbo2019learning, peng2018deepmimic}, which are often tricky to design by humans due to heavy domain expertise requirements and tedious trials and errors. 
In contrast, sparse rewards, such as a binary task completion signal, are significantly easier to obtain (often directly from the environment).
For instance, in pick-and-place tasks, the sparse reward could simply be defined as the object being placed at the goal location. Nonetheless, sparse rewards also introduce challenges (e.g., exploration) for RL algorithms~\citep{pathak2017curiosity, rnd, ecoffet2019go}. Therefore, a crucial question arises: \emph{can we learn dense reward functions in a data-driven manner?}

Ideally, the learned reward will be \textbf{reused} to efficiently solve new tasks that share similar success conditions with the task used to learn the reward. For example, in pick-and-place tasks, different objects may need to be manipulated with varying dynamics, action spaces, and even robot morphologies. For clarity, we refer to each variant as a \emph{task} and the set of all possible pick-and-place tasks as a \emph{task family}. Importantly, the reward function, which captures approaching, grasping, and moving the object toward the goal position, can potentially be transferred within this task family. 
This observation motivates us to explore the concept of \emph{reusable rewards}, which can be learned as a function from some tasks and reused in unseen tasks. While existing literature in RL primarily focuses on the reusability (generalizability) of policies, we argue that rewards can pose greater flexibility for reuse across tasks. 
For example, it is nearly impossible to directly transfer a policy operating a two-finger gripper for pick-and-place to a three-finger gripper due to action space misalignment, but a reward inducing the approach-grasp-move workflow may apply for both types of grippers.

However, many existing works on reward learning do not emphasize reward reuse for new tasks. The field of learning a reward function from demonstrations is known as inverse RL in the literature~\citep{algo_irl, abbeel2004apprenticeship, max_entropy_irl}. More recently, adversarial imitation learning (AIL) approaches have been proposed \citep{gail, dac, airl, ghasemipour2020divergence} and gained popularity. Following the paradigm of GANs~\citep{gan}, AIL approaches employ a policy network to generate trajectories and train a discriminator to distinguish between agent trajectories from demonstration ones. 
By using the discriminator score as rewards, \citep{gail} shows that a policy can be trained to imitate the demonstrations. 
Unfortunately, such rewards are \textit{not reusable} across tasks -- at convergence, the discriminator outputs $\frac{1}{2}$ for both the agent trajectories and the demonstrations, as discussed in \citep{gan, airl}, making it unable to learn useful information for solving new tasks.

In contrast to AIL, we propose a novel approach for learning \textit{reusable rewards}. Our approach involves incorporating sparse rewards as a supervision signal in lieu of the original signal used for classifying demonstration and agent trajectories. Specifically, we train a discriminator to classify \textit{success trajectories} and \textit{failure trajectories} based on the binary sparse reward. Please refer to Fig.~\ref{fig:method} (a)(b) for an illustrative depiction. 
Our formulation assigns higher rewards to transitions in success trajectories and lower rewards to transitions within failure trajectories, which is consistent throughout the entire training process. 
As a result, the reward will be reusable once the training is completed. 
Expert demonstrations can be included as success trajectories in our approach, though they are not mandatory. We only require the availability of a sparse reward, which is a relatively weak requirement as it is often an inherent component of the task definition.

\begin{figure*}[t]
    \centering
    \vspace{-1cm}
    \includegraphics[width=\linewidth]{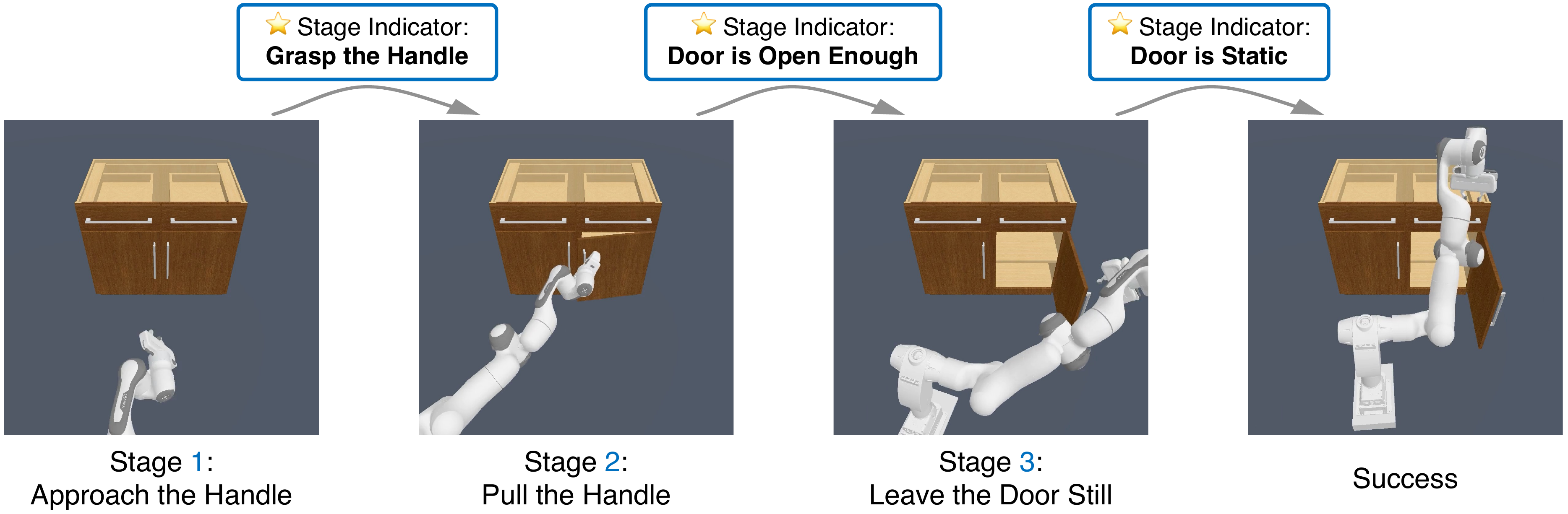}
    \vspace{-0.4cm}
    \caption{An illustration of stage indicators in an OpenCabinetDoor task, which can be naturally divided into three stages 
    plus a success state. A stage indicator is a binary function representing whether the current state is in a certain stage, and it can be simply defined by some boolean functions.
    }
    \label{fig:stage_illustration}
    \vspace{-0.4 cm}
\end{figure*}

Our approach can be extended to leverage the inherent structure of \textit{multi-stage tasks} and derive stronger dense rewards. Many tasks naturally exhibit multi-stage structures, and it is relatively easy to assign a binary indicator on whether the agent has entered a stage. For example, in the "Open Cabinet Door" task depicted in Fig.~\ref{fig:stage_illustration}, there are three stages: 1) approach the door handle, 2) grasp the handle and pull the door, and 3) release the handle and keeping it steady.
If the agent is grasping the handle of the door but the door has not been opened enough, then we can simply use a corresponding binary indicator asserting that the agent is in the 2nd stage.
\footnote{Stage indicators are only required during RL training, but \textbf{not required} when deploying policy to real world.}
By utilizing these stage indicators, we can learn a dense reward for each stage and combine them into a more structured reward. Since the horizon for each stage is shorter than that of the entire task, learning a high-quality dense reward becomes more feasible.
Furthermore, this approach provides flexibility in incorporating extra information beyond the final success signal. 
We dub our approach as \textbf{DrS} (\textbf{D}ense \textbf{r}eward learning from \textbf{S}tages).

Our approach exhibits strong performance on challenging tasks. To assess the reusability of the rewards learned by our approach, we employ the ManiSkill benchmark~\citep{mu2021maniskill, gu2023maniskill2}, which offers a large number of task variants within each task family.We evaluate our approach on three task families: Pick-and-Place, Open Cabinet Door, and Turn Faucet, including 1000+ task variants. 
Each task variant involves manipulating a different object and requires precise low-level physical control, thereby highlighting the need for a good dense reward. 
Our results demonstrate that the learned rewards can be reused across tasks, leading to improved performance and sample efficiency of RL algorithms compared to using sparse rewards. In certain tasks, the learned rewards even achieve performance comparable to those attained by human-engineered reward functions.

Moreover, our approach \textbf{drastically reduces the human effort} needed for reward engineering. For instance, while the human-engineered reward for "Open Cabinet Door" involves \textit{over 100 lines of code, 10 candidate terms, and tons of "magic" parameters}, our approach only requires \textit{two boolean functions} as stage indicators: if the robot has grasped the handle and if the door is open enough. See appendix \ref{sec:human_effort_comparison} for a detailed example illustrating how our method reduces the required human effort.

Our contributions can be summarized as follows: 
\setlist*[itemize]{labelindent=\parindent,itemindent=0pt,leftmargin=20pt}
\begin{itemize}%
    \item We propose \textbf{DrS} (\textbf{D}ense \textbf{r}eward learning from \textbf{S}tages), a novel approach for learning reusable dense rewards for multi-stage tasks, effectively reducing human efforts in reward engineering.
    \item Extensive experiments on 1,000+ task variants from three task families 
    showcase the effectiveness of our approach in generating high-quality and reusable dense rewards.
\end{itemize}

\vspace{-0.2cm}
\section{Related Works}
\vspace{-0.2cm}

\noindent \textbf{Learning Reward from Demo (Offline)}
Designing rewards is challenging due to domain knowledge requirements, so approaches to learning rewards from data have gained attention. 
Some methods adopt classification-based rewards, i.e., training a reward by classifying goals \citep{smith2019avid, mt_opt, du2023vision} or demonstration trajectories \citep{oril}. Other methods \citep{xirl, aytar2018playing} use the distance to goal as a reward function, where the distance is usually computed in a learned embedding space, but these methods usually require that the goal never changes in a task.
These rewards are only trained on offline datasets, hence they can \textit{easily be exploited} by an RL agent, i.e., an RL can enter a state that is not in the dataset and get a wrong reward signal, as studied in \citep{vecerik2019practical, purl}.

\noindent \textbf{Learning Reward from Demo (Online)}
The above issue can be addressed by allowing agents to verify the reward in the environment, and inverse reinforcement learning (IRL) is the prominent paradigm. 
IRL aims to recover a reward function given expert demonstrations. 
Traditional IRL methods \citep{algo_irl, abbeel2004apprenticeship, max_entropy_irl, ratliff2006maximum} often require multiple iterations of Markov Decision Process solvers \citep{mdp_book}, resulting in poor sample efficiency.
In recent years, adversarial imitation learning (AIL) approaches are proposed
\citep{gail, dac, airl, ghasemipour2020divergence, liu2019state}. 
They operate similarly to generative adversarial networks (GANs) \citep{gan}, in which a generator (the policy) is trained to maximize the confusion of a discriminator, and the discriminator (serves the role of rewards) is trained to classify the agent trajectories and demonstrations. 
However, such rewards are \textit{not reusable} as we discussed in the introduction - classifying agent trajectories and demonstrations is impossible at convergence.
In contrast, our approach gets rid of this issue by classifying the success/failure trajectories instead of expert/agent trajectories.

\noindent \textbf{Learning Reward from Human Feedback~}
Recent studies \citep{christiano2017deep, ibarz2018reward, abc} infer the reward through human preference queries on trajectories or explicitly asking for trajectory rankings \citep{brown2019extrapolating}. Another line of works \citep{vice, raq} involves humans specifying desired outcomes or goals to learn rewards. 
However, in these methods, the rewards only distinguish goal from non-goal states, offering relatively weak incentives to agents at the beginning of an episode, especially in long-horizon tasks. In contrast, our approach classifies all the states in the trajectories, providing strong guidance throughout the entire episode.

\noindent \textbf{Reward Shaping~}
Reward shaping methods aim to densify sparse rewards. 
Earlier works \citep{ng1999policy} study the forms of shaped rewards that induce the same optimal policy as the ground-truth reward. 
Recently, some works \citep{trott2019keeping, drem} have shaped the rewards as the distance to the goal, similar to some offline reward learning methods mentioned above. 
Another idea \citep{memarian2021self} involves shaping delayed reward by ranking trajectories based on a fine-grained preference oracle. 
In contrast to these reward shaping approaches, our method leverages demonstrations, which are available in many real-world problems \citep{sun2020scalability, dasari2019robonet}. This not only boosts the reward learning process but also reduces the additional domain knowledge required by these methods.

\noindent \textbf{Task Decomposition~} The decomposition of tasks into stages/sub-tasks has been explored in various domains. Hierarchical RL approaches \citep{
frans2018meta, nachum2018data, levy2018learning} break down policies into sub-policies to solve specific sub-tasks. Skill chaining methods \citep{lee2021adversarial, gu2022multi, lee2019composing} focus on solving long-horizon tasks by combining multiple short-horizon policies or skills. Recently, language models have also been utilized to break the whole task into sub-tasks \cite{saycan}. In contrast to these approaches that utilize stage structures in policy space, our work explores an orthogonal direction by designing rewards with stage structures.

\section{Problem Setup}

\begin{figure*}[t]
    \centering
    \vspace{-0.8cm}
    \includegraphics[width=\linewidth]{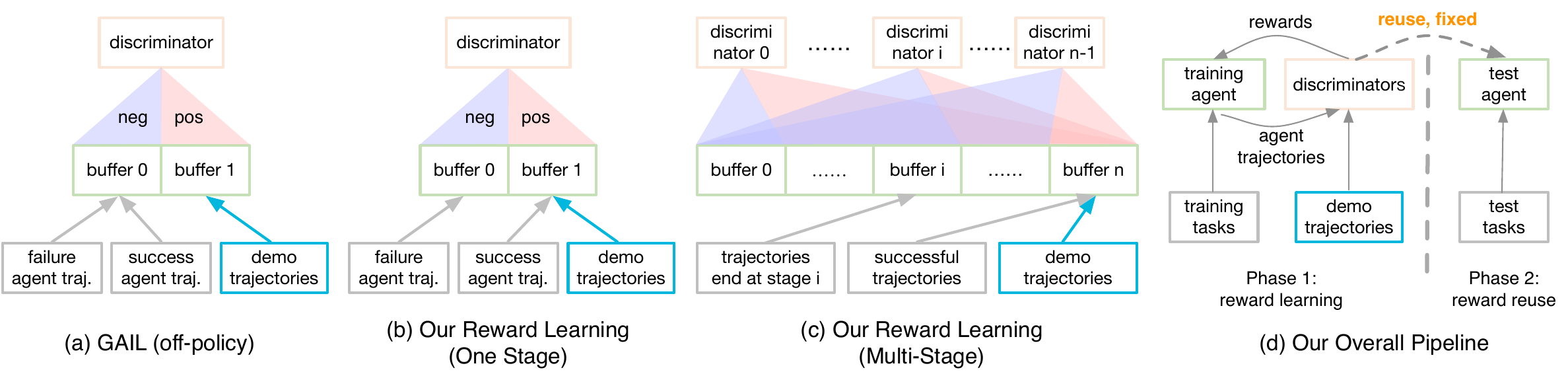}
    \vspace{-0.6cm}
    \caption{ %
a) GAIL's discriminator aims to distinguish agent trajectories from demonstrations.
b) In single-stage tasks, the discriminator in our approach aims to distinguish success trajectories from failure ones.
c) In multi-stage tasks, our approach train a separate discriminator for each stage. The discriminator for stage $k$ aims to distinguish trajectories that reach beyond stage $k$ from those that only reach up to stage $k$.
d) Overall, our approach has 2 phases: \textbf{reward learning} and \textbf{reward reuse}.
    }
    \vspace{-0.4cm}
    \label{fig:method}
    
\end{figure*}

In this work, we adopt the Markov Decision Process (MDP) $\mathcal{M}:=\langle S,A,T,R,\gamma \rangle$ as the theoretical framework, where $R$ is a reward function that defines the goal or purpose of a task. Specifically, we focus on tasks with \textit{sparse rewards}. In this context, "sparse reward" denotes a binary reward function that gives a value of $1$ upon successful task completion and $0$ otherwise:

\vspace{-0.5cm}
\begin{equation}
    R_{sparse}(s) = 
    \begin{cases}
        ~1 & \text{task is completed by reaching one of the success states } s \\
        ~0 & \text{otherwise}
    \end{cases}
\end{equation}
\vspace{-0.2cm}

Our objective is to learn a dense reward function from a set of training tasks, with the intention of reusing it for unseen test tasks. 
Specifically, we aim to successfully train RL agents from scratch on the test tasks using the learned rewards. The desired outcome is to enhance the efficiency of RL training, surpassing the performance achieved by sparse rewards.

We assume that both the training and test tasks are in the same \emph{task family}. A task family refers to a set of task variants that share the same success criteria, but may differ in terms of assets, initial states, transition functions, and other factors. For instance, the task family of object grasping includes tasks such as "Alice robot grasps an apple" and "Bob robot grasps a pen." The key point is that tasks within the same task family share a common underlying sparse reward.

Additionally, we posit that the task can be segmented into multiple stages, and the agent has access to several \emph{stage indicators} obtained from the environment. 
A stage indicator is a binary function that indicates whether the current state corresponds to a specific stage of the task. An example of stage indicators is in Fig.~\ref{fig:stage_illustration}. 
This assumption is quite general as many long-term tasks have multi-stage structures, and determining the current stage of the task is not hard in many cases. %
By utilizing these stage indicators, it becomes possible to construct a reward that is slightly denser than the binary sparse reward, which we refer to as a \emph{semi-sparse} reward, and it serves as a strong baseline:
\begin{equation}
    R_{semi-sparse}(s) = k \text{, when state $s$ is at stage }k
    \label{eq:semi-sparse}
\end{equation}

We aim to design an approach that learns a dense reward based on the stage indicators. When expert demonstration trajectories are available, they can also be incorporated to boost the learning process. 

Note that the stage indicators are only required during RL training, but \textit{not required when deploying the policy to the real world}. Training RL agents directly in the real world is often impractical due to cost and safety issues. Instead, a more common practice is to train the agent in simulators and then transfer/deploy it to the real world. 
While obtaining the stage indicators in simulators is fairly easy, it is also possible to obtain them in the real world by various techniques (robot proprioception, tactile sensors \cite{lin2022tactile, melnik2021using}, visual detection/tracking \cite{qt_opt, mt_opt}, large vision-language models \cite{du2023vision}, etc.).

\section{DrS: Dense reward learning from Stages}

Dense rewards are often tricky to design by humans (see an example in appendix \ref{sec:human_effort_comparison}), so
we aim to learn a reusable dense reward function from stage indicators in multi-stage tasks and demonstrations when available. Overall, our approach has two phases, as shown in Fig.~\ref{fig:method} (d): 
\begin{itemize}
    \item \textbf{Reward Learning Phase}: learn the dense reward function using training tasks.
    \item \textbf{Reward Reuse Phase}: reuse the learned dense reward to train new RL agents in test tasks.
\end{itemize}

Since the reward reuse phase is just a regular RL training process, we only discuss the reward learning phase in this section. 
We first explain how our approach learns a dense reward in one-stage tasks (Sec. \ref{sec:method_one_stage}). Then, we extend this approach to multi-stage tasks (Sec. \ref{sec:method_multi_stage}). 

\subsection{Reward Learning on One-Stage Tasks}
\label{sec:method_one_stage}

In line with previous work \citep{vecerik2019practical, vice}, we employ a classification-based dense reward. We train a classifier to distinguish between good and bad trajectories, utilizing the learned classifier as dense reward. Essentially, states resembling those in good trajectories receive higher rewards, while states resembling bad trajectories receive lower rewards.
While previous Adversarial Imitation Learning (AIL) methods 
\citep{gail, dac} 
used discriminators as classifiers/rewards to distinguish between agent and demonstration trajectories, these discriminators cannot be directly \textit{reused} as rewards to train new RL agents. As the policy improves, the agent trajectories (negative data) and the demonstrations (positive data) can become nearly identical. Therefore, at convergence, the discriminator output for both agent trajectories and demonstrations tends to approach $\frac{1}{2}$, as observed in GANs \citep{gan} (also noted by \citep{airl, purl}). This makes it unable to learn useful info for solving new tasks.

Our approach introduces a simple modification to existing AIL methods to ensure that the discriminator continues to learn meaningful information even at convergence. The key issue previously mentioned arises from the diminishing gap between agent and demonstration trajectories over time, making it challenging to differentiate between positive and negative data. To address this, we propose training the discriminator to distinguish between success and failure trajectories instead of agent and demonstration trajectories. By defining success and failure trajectories based on the sparse reward signal from the environment, the gap between them remains intact and does not shrink. 
Consequently, the discriminator effectively emulates the sparse reward signal, providing dense reward signals to the RL agent. Intuitively, a state that is closer to the success states in terms of task progress (rather than Euclidean distance) receives a higher reward, as it is more likely to occur in success trajectories. Fig.~\ref{fig:method}(a) and (b) illustrate the distinction between our approach and traditional AIL methods.

To ensure that the training data consistently includes both success and failure trajectories, we use replay buffers to store historical experiences, and train the discriminator in an off-policy manner. While the original GAIL is on-policy, recent AIL methods \citep{dac, orsini2021matters} have adopted off-policy training for better sample efficiency.
Note that although our approach shares similarities with AIL methods, it is not adversarial in nature. In particular, our policy does not aim to deceive the discriminator, and the discriminator does not seek to penalize the agent's trajectories.

\begin{wrapfigure}{r}{0.33\linewidth}
    \centering
    \vspace{-1.3cm}
    \includegraphics[width=\linewidth]{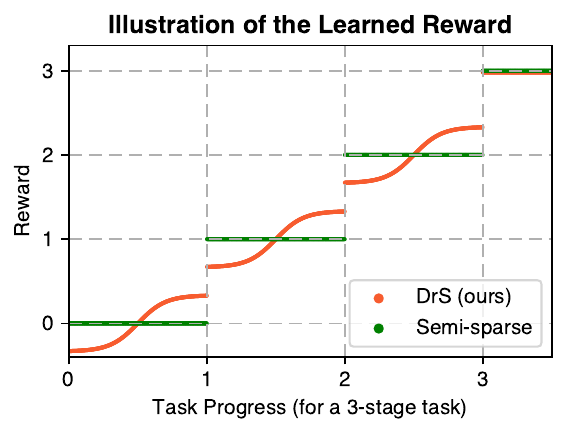}
    \vspace{-0.7cm}
    \caption{An illustration of our learned reward, which fills the gaps in semi-sparse rewards, resulting in a smooth reward curve.}
    \label{fig:reward_illustration}
    \vspace{-0.2cm}
\end{wrapfigure}

\subsection{Reward Learning on Multi-Stage Tasks}
\label{sec:method_multi_stage}

In multi-stage tasks, it is desirable for the reward of a state in stage $k+1$ to be strictly higher than that of stage $k$ to incentivize the agent to progress towards later stages. The semi-sparse reward (Eq.~\ref{eq:semi-sparse}) aligns with this intuition, but it is still a bit too sparse. If each stage of the task is viewed as an individual task, the semi-sparse reward acts as a sparse reward for each stage. In the case of a one-stage task, a discriminator can be employed to provide a dense reward. 
Similarly, for multi-stage tasks, a separate discriminator can be trained for each stage to serve as a dense reward for that particular stage. By training stage-specific discriminators, we can effectively address the sparse reward issue and guide the agent's progress through the different stages of the task.
Fig. \ref{fig:reward_illustration} gives an intuitive illustration of our learned reward, which fills the gaps in semi-sparse rewards, resulting in a smooth reward curve.

To train the discriminators for different stages, we need to establish the positive and negative data for each discriminator. In one-stage tasks, positive data comprises success trajectories and negative data encompasses failure trajectories. 
In multi-stage tasks, we adopt a similar approach with a slight modification. Specifically, we assign a stage index to each trajectory, which is determined as the highest stage index among all states within the trajectory: 

\vspace{-0.4cm}
\begin{equation}
    \text{StageIndex}(\tau: (s_0, s_1, ...))=\max_i ~\text{StageIndex}(s_i),
    \label{eq:stage_index}
\end{equation}
where $\tau$ is a trajectory and $s_i$ are the states in $\tau$. For the discriminator associated with stage $k$, positive data consists of trajectories that progress beyond stage $k$ (StageIndex $>k$), and negative data consists of trajectories that reach up to stage $k$ (StageIndex $\le k$).

Once the positive and negative data for each discriminator have been established, the next step is to combine these discriminators to create a reward function. While the semi-sparse reward (Eq.~\ref{eq:semi-sparse}) lacks incentives for the agent at stage $k$ until it reaches stage $k+1$, we can fill in the gaps in the semi-sparse reward by the stage-specific discriminators. We define our learned reward function for a multi-stage task as follows:

\vspace{-0.4cm}
\begin{equation}
R(s') = k + \alpha \cdot \tanh(\text{Discriminator}_k(s'))
\label{equ:reward}
\end{equation}
where $k$ is the stage index of $s'$ and $\alpha$ is a hyperparameter. Basically, the formula incorporates a dense reward term into the semi-sparse reward. The $\tanh$ function is used to bound the output of the discriminators. As the range of the $\tanh$ function is (-1, 1), any $\alpha<\frac{1}{2}$ ensures that the reward of a state in stage $k+1$ is always higher than that of stage $k$. In practice, we use $\alpha=\frac{1}{3}$ and it works well.

\newcommand{\data}{\mathcal{D}}
\newcommand{\obs}{\mathbf{s}}
\newcommand{\state}{\mathbf{s}}
\newcommand{\act}{\mathbf{a}}
\newcommand{\posdata}{\data^+}
\newcommand{\loss}{\mathcal{L}}
\newcommand{\out}{y}
\newcommand{\policy}{\pi}

\begin{wrapfigure}{R}{0.6\textwidth}
\begin{minipage}{0.6\textwidth}
\vspace{-1.7cm}

\begin{algorithm}[H]
    \caption{\textbf{DrS} (\textbf{D}ense \textbf{r}eward learning from \textbf{S}tages )}
    \small
    \label{alg:ours}
    \begin{algorithmic}[1]
    \Require Task MDP $\mathcal{M}$, Number of stages in task $N$, Demonstration dataset $\data := \{\tau^0, \tau^1, ...  \}$ (optional)
    \State Initialize policy $\policy$, critic $Q$, replay buffer $\mathcal{B}_R$
    \State Initialize discriminators $f_0, f_1, ..., f_{N-1}$, stage buffers $\mathcal{B}_0, \mathcal{B}_2, .., \mathcal{B}_N$
    \State Fill demo $\data$ into $\mathcal{B_N}$: $\mathcal{B_N}\leftarrow \mathcal{B_N} \cup \data$
    \For {each iteration} 
    \State Collect trajectories $\{\tau_\pi^0, \tau_\pi^1, ...\}$ by executing $\pi$ in $\mathcal{M}$
    \State Add trajectories to replay buffer: $\mathcal{B}_R \leftarrow \mathcal{B}_R \cup \{\tau_\pi^0, \tau_\pi^1, ...\}$
    \For {each trajectory $\tau_\pi^i$ in $\{\tau_\pi^0, \tau_\pi^1, ...\}$}
        \State $j=\text{StageIndex}(\tau_\pi^i)$ according to Eq. \ref{eq:stage_index}
        \State $\mathcal{B}_j \leftarrow \mathcal{B}_j \cup \{\tau_\pi^i\}$
    \EndFor
    \For {each gradient step for discriminators}
        \For {each discriminator $f_k$}
        \State Sample negative data from $\bigcup_{i=0}^k \mathcal{B}_i$
        \State Sample positive data from $\bigcup_{i=k+1}^N \mathcal{B}_i$
        \State Update $f_k$ using BCE loss 
        \EndFor
    \EndFor

    \For {each gradient step for the policy $\policy$}
    \State Sample from $\mathcal{B}_R$
    \State Compute rewards according to Eq. \ref{equ:reward}
    \State Update $\policy$ and $Q$ by SAC~\citep{sac}
    
    \EndFor

    \EndFor
    \end{algorithmic}
\end{algorithm}
\vspace{-1 cm}

\end{minipage}
\end{wrapfigure}

\subsection{Implementation}
\label{sec:implementation}
From the implementation perspective, our approach is similar to GAIL, but with a different training process for discriminators. While the original GAIL is combined with TRPO \citep{trpo}, \citep{orsini2021matters} found that using state-of-the-art off-policy RL algorithms (like SAC \citep{sac} or TD3 \citep{td3}) can greatly improve the sample efficiency of GAIL. Therefore, we also combine our approach with SAC, and the full algorithm is summarized in Algo. \ref{alg:ours}.

In addition to the regular replay buffer used in SAC, our approach maintains $N$ different stage buffers to store trajectories corresponding to different stages(defined by Eq. \ref{eq:stage_index}). Each trajectory is assigned to only one stage buffer based on its stage index. During the training of the discriminators, we sample data from the union of multiple buffers.
In practice, we early stop the discriminator training of $k$ once its success rate is sufficiently high, as we find it reduces the computational cost and makes the learned reward more robust. 
Note that our approach uses the next state $s'$ as the input to the reward, which aligns with common practices in human reward engineering ~\citep{gu2023maniskill2, zhu2020robosuite}. However, our approach is also compatible with alternative forms of input, such as $(s,a)$ or $(s, a, s')$.

\section{Experiments}
\label{sec:exp}

\subsection{Setup and Task Descriptions}

\begin{figure*}[t]
    \centering
    \vspace{-1 cm}
    \includegraphics[width=\linewidth]{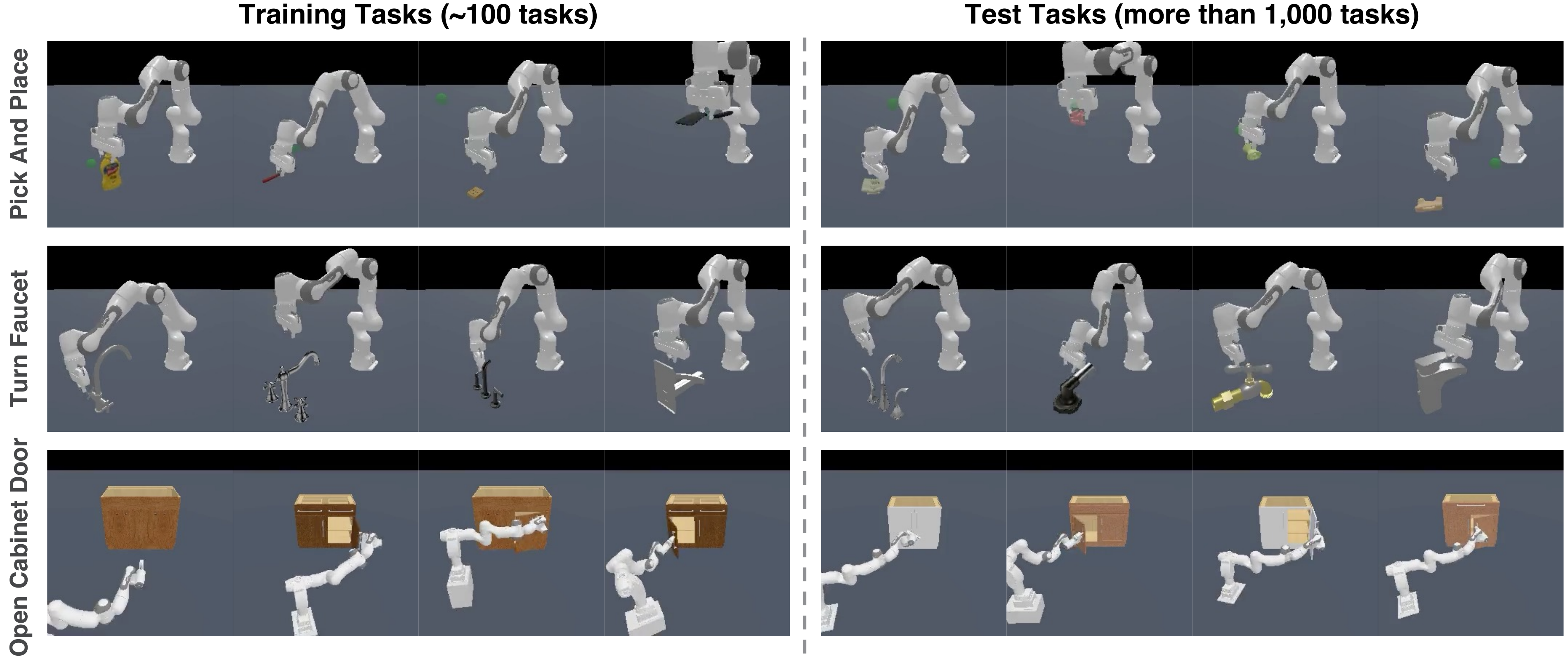}
    \vspace{-0.5 cm}
    \caption{We evaluated our approach DrS on more than 1,000 task variants from three task families in ManiSkill~\citep{mu2021maniskill, gu2023maniskill2}. Each task variant is associated with a different object. All tasks require low-level physical control. The objects in training and test tasks are non-overlapped. 
    }
    \label{fig:object_variation}
    \vspace{-0.5 cm}
\end{figure*}

We evaluated our approach on three challenging physical manipulation task families from the ManiSkill~\citep{mu2021maniskill, gu2023maniskill2}: Pick-and-Place, Turn Faucet, and Open Cabinet Door. 
Each task family includes a set of different objects to be manipulated. To assess the reusability of the learned rewards, we divided the objects within each task family into non-overlapping training and test sets, as depicted in Fig.\ref{fig:object_variation}.
During the reward learning phase, we learned the rewards by training an agent for each task family to manipulate all training objects. 
In the subsequent reward reuse phase, the learned reward rewards are reused to train an agent to manipulate all test objects for each task family. 
And we compare with other baseline rewards in this \textbf{reward reuse} phase.
It is important to note that our learned rewards are agnostic to the specific RL algorithm employed. However, we utilized the Soft Actor-Critic (SAC) algorithm to evaluate the quality of the different rewards.

To assess the \textit{reusability} of the learned rewards, it is crucial to have a diverse set of tasks that exhibit similar structures and goals but possess variations in other aspects. However, most existing benchmarks lack an adequate number of task variations within the same task family. As a result, we primarily conducted our evaluation on the ManiSkill benchmark, which offers a range of object variations within each task family. 
This allowed us to thoroughly evaluate our learned rewards in a realistic and comprehensive manner. %

\noindent \textbf{Pick-and-Place:} A robot arm is tasked with picking up an object and relocating it to a random goal position in mid-air. The task is completed if the object is in close proximity to the goal position, and both the robot arm and the object remain stationary. 
The stage indicators include: (a) the gripper grasps the object, (b) the object is close the goal position, and (c) both the robot and the object are stationary. We learn rewards on 74 YCB objects and reuse rewards on 1,600 EGAD objects.

\noindent \textbf{Turn Faucet:} A robot arm is tasked to turn on a faucet by rotating its handle. The task is completed if the handle reaches a target angle. The stage indicators include: (a) the target handle starts moving, (b) the handle reaches a target angle. We learn rewards on 10 faucets and reuse rewards on 50 faucets.

\noindent \textbf{Open Cabinet Door:} A single-arm mobile robot is required to open a designated target door on a cabinet. The task is completed if the target door is opened to a sufficient degree and remains stationary. The stage indicators include: (a) the robot grasps the door handle, (b) the door is open enough, and (c) the door is stationary. We learn rewards on 4 cabinet doors and reuse rewards on 6 cabinet doors. Note that we remove all single-door cabinets in this task family, as they can be solved by kicking the side of the door and this behavior can be readily learned by sparse rewards.

We employed low-level physical control for all task families. Please refer to the appendix \ref{sec:task_desc} for a detailed description of the object sets, action space, state space, and demonstration trajectories.

\vspace{-0.2cm}
\subsection{Baselines}
\vspace{-0.2cm}

\noindent \textbf{Human-Engineered~} The original human-written dense rewards in the benchmark, which require a significant amount of domain knowledge, thus can be considered as an \textit{upper bound} of performance.

\noindent \textbf{Semi-Sparse~} The rewards constructed based on the stage indicators, as discussed in Eq.~\ref{eq:semi-sparse}. The agent receives a reward of $k$ when it is in stage $k$. This baseline extends the binary sparse reward.

\noindent \textbf{VICE-RAQ~\citep{raq}} An improved version of VICE~\citep{vice}. It learns a classifier, where the positive samples are successful states annotated by querying humans, and the negative samples are all other states collected by the agent. Since our experiments do not involve human feedback, we let VICE-RAQ query the oracle success condition infinitely for a fair comparison.

\noindent \textbf{ORIL~\citep{oril}} A representative offline reward learning method, where the agent does not interact with the environments but purely learns from the demonstrations.
It learns a classifier (reward) to distinguish between the states from success trajectories and random trajectories.

\subsection{Comparison with Baseline Rewards}

\begin{figure*}[t]
    \centering
    \vspace{-0.8cm}
    \includegraphics[width=\linewidth]{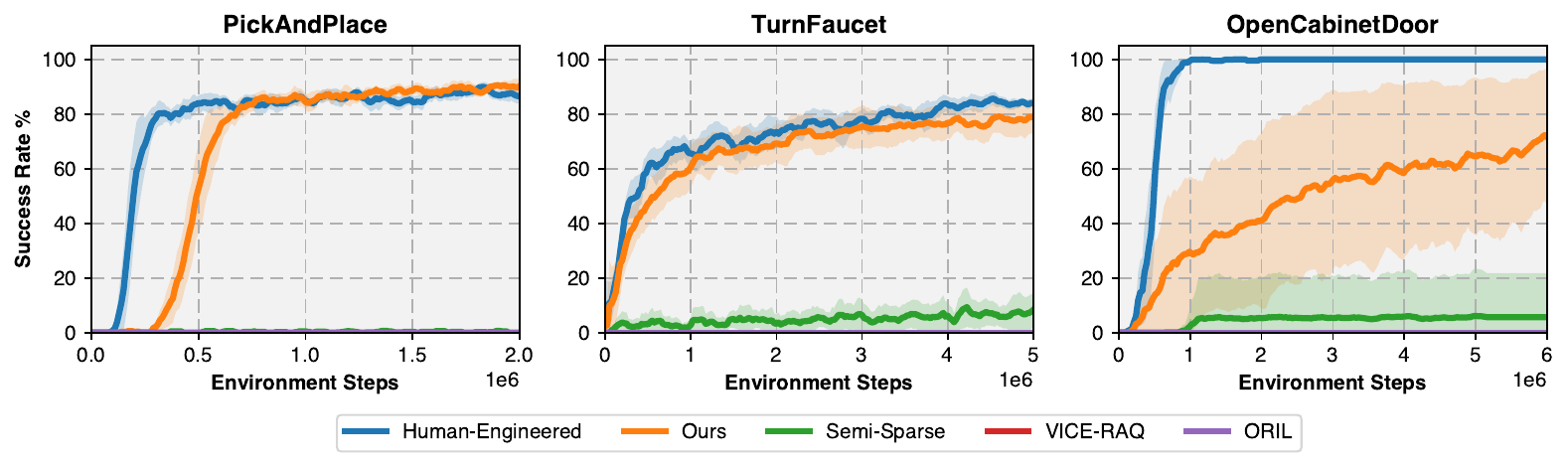}
    \vspace{-0.5cm}
    \caption{Evaluation results of \textbf{reusing learned rewards}. All curves use SAC to train, but with different rewards. VICE-RAQ and ORIL get no success. 5 random seeds, the shaded region is std.}
    \label{fig:main_results}
    \vspace{-0.5cm}
\end{figure*}

We trained RL agents using various rewards and assessed the reward quality based on both the sample efficiency and final performance of the agents. 
The experimental results, depicted in Fig.~\ref{fig:main_results}, demonstrate that our learned reward surpasses semi-sparse rewards and all other reward learning methods across all three task families. 
This outcome suggests that our approach successfully acquires high-quality rewards that significantly enhance RL training. 
Remarkably, our learned rewards even achieve performance comparable to human-engineered rewards in Pick-and-Place and Turn Faucet.

Semi-sparse rewards yielded limited success within the allocated training budget, suggesting that RL agents face exploration challenges when confronted with sparse reward signals.
VICE-RAQ failed in all tasks. Notably, it actually failed during the reward learning phase on the training tasks, rendering the learned rewards inadequate for supporting RL training on the test tasks. 
This failure aligns with observations made by \citep{drem}. We hypothesize that by only classifying the success states from other states, it cannot provide sufficient guidance during the early stages of training, where most states are distant from the success states and receive low rewards.
Unsurprisingly, ORIL does not get any success on all tasks either. Without interacting with the environments to gather more data, the learned reward functions easily tend to overfit the provided dataset.
When using such rewards in RL, the flaws in the learned rewards are easily exploited by the RL agents.

\subsection{Ablation Study}

We examined various design choices within our approach on the Pick-and-Place task family.

\subsubsection{Robustness to Stage Configurations}

Though many tasks present a natural structure of stages, there are still different ways to divide a task into stages. To assess the robustness of our approach in handling different task structures, we experiment with different numbers of stages and different ways to define stage indicators.

\begin{figure}[t]
    \vspace{-0.4 cm}
    \centering
    \begin{minipage}{.31\linewidth}
        \centering
        \includegraphics[width=1.1\linewidth]{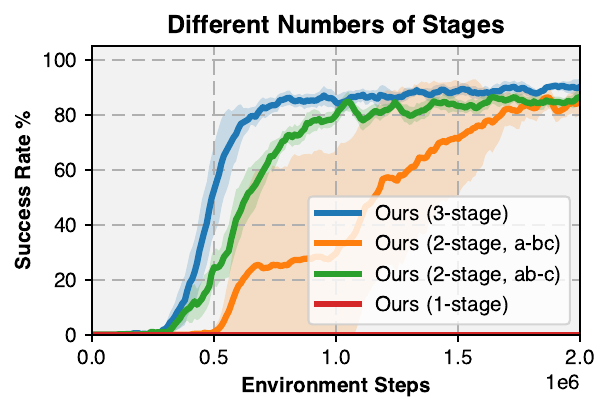}
        \vspace{-0.5 cm}
        \caption{
        Ablation study on the number of stages, see \hyperref[par:num_stages]{here}.
        }
        \label{fig:ablation_num_stages}
    \end{minipage}
    \hspace{0.01\linewidth} 
    \begin{minipage}{.31\linewidth}
        \centering
        \includegraphics[width=1.1\linewidth]{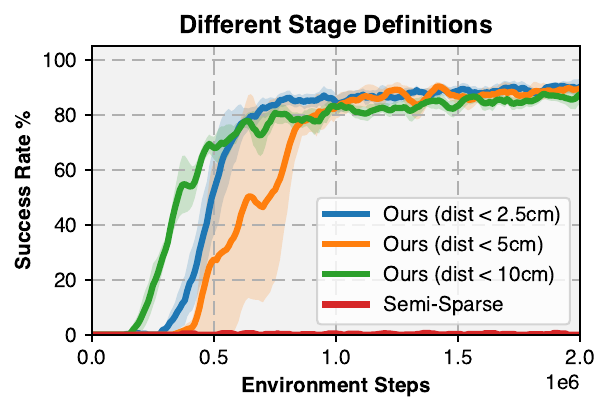}
        \vspace{-0.5 cm}
        \caption{
            Ablation study on the stage definitions, see \hyperref[par:def_stages]{here}.
        }
        \label{fig:ablation_stage_definition}
    \end{minipage}
    \hspace{0.01\linewidth} 
    \begin{minipage}{.31\linewidth}
        \centering
        \includegraphics[width=1.1\linewidth]{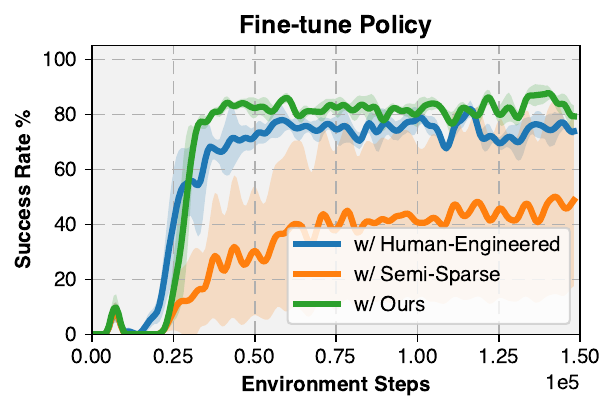}
        \vspace{-0.5 cm}
        \caption{
        Fine-tune the policy from reward learning, see \hyperref[sec:ft_policy]{here}. 
        }
        \label{fig:finetune}
    \end{minipage}
    \vspace{-0.4 cm}
\end{figure}

\paragraph{Number of Stages} \label{par:num_stages}
The Pick-and-Place task family originally consisted of three stages: (a) approach the object, (b) move the object to the goal, and (c) make everything stationary. We explored two ways of reducing the number of stages to two, namely merging stages (a) and (b) or merging stages (b) and (c), as well as the 1-stage case. 
Our results, presented in Fig. \ref{fig:ablation_num_stages}, indicate that the learned rewards with 2 stages can still effectively train RL agents in test tasks, albeit with lower sample efficiency than those with 3 stages. 
Specifically, the reward that preserves stage (c) "make everything stationary" performs slightly better than the reward that preserves stage (a) "approach the object". This suggests that it may be more challenging for a robot to learn to stop abruptly without a dedicated stage. 
However, when reducing the number of stages to 1, the learned reward failed to train RL agents in test tasks, demonstrating the benefit of using more stages in our approach.

\paragraph{Definition of Stages} \label{par:def_stages}
The stage indicator "object is placed" is initially defined as if the distance between the object and the goal is less than 2.5 cm. We create two variants of it, where the distance thresholds are 5cm and 10cm, respectively. The results, as depicted in Fig. \ref{fig:ablation_stage_definition}, demonstrate that changing the distance threshold within a reasonable range does not significantly affect the efficiency of RL training. Note that the task success condition is unchanged, and our rewards consistently encourage the agents to reach the success state as it yields the highest reward according to Eq. \ref{equ:reward}. The stage definitions solely affect the efficiency of RL training during the reward reuse phase.

Overall, the above results highlight the robustness of our approach to different stage configurations, indicating that it is not heavily reliant on intricate stage designs. This robustness contributes to a significant reduction in the burden of human reward engineering.

\subsubsection{Fine-tuning Policy}
\label{sec:ft_policy}

In our previous experiments, we assessed the quality of the learned reward by reusing it in training RL agents from scratch since it is the most common and natural way to use a reward. 
However, our approach also produces a policy as a byproduct in the reward learning phase. This policy can also be fine-tuned using various rewards in new tasks, providing an alternative to training RL agents from scratch. We compare the fine-tuning of the byproduct policy using human-engineered rewards, semi-sparse rewards, and our learned rewards.

As shown in Fig.~\ref{fig:finetune}, all policies improve rapidly at the beginning due to the good initialization of the policies. However, fine-tuning with our learned reward yields the best performance (even slightly better than the human-engineered reward), indicating the advantages of utilizing our learned dense reward even with a good initialization. 
Furthermore, the significant variance observed when fine-tuning the policy with semi-sparse rewards highlights the limitations of sparse reward signals in effectively training RL agents, even with a very good initialization.

\subsubsection{Additional Ablation Studies}
Additional ablation studies are provided in appendix \ref{sec:additional_ablation}, with key conclusions summarized as follows:
\vspace{-0.2cm}
\begin{itemize}
    \item DrS is compatible with various modalities of reward input, including point cloud data. \ref{sec:pcd} 
    \item Reward learned by GAIL, even with stage indicators, is not reusable. \ref{sec:basic_ablation}
    \item The way of combining the dense rewards from each stage matters. \ref{sec:reward_formula}
\end{itemize}

\vspace{-0.2cm}

\section{Conclusion and Limitations}
\vspace{-0.2cm}

To make RL a more widely applicable tool, we have developed a data-driven approach for learning dense reward functions that can be reused in new tasks from sparse rewards. We have evaluated the effectiveness of our approach on robotic manipulation tasks, which have high-dimensional action spaces and require dense rewards. Our results indicate that the learned dense rewards are effective in transferring across tasks with significant variation in object geometry. 
By simplifying the reward design process, our approach paves the way for scaling up RL in diverse scenarios.

We would like to discuss two main limitations when using the multi-stage version of our approach. 

Firstly, though our experiments show the substantial benefits of knowing the multi-stage structure of tasks (at training time, not needed at policy deployment time), we did not specifically investigate how this knowledge can be acquired. Much future work on be done here, by leveraging large language models such as ChatGPT~\citep{openai2023gpt4} (by our testing, they suggest stages highly aligned to the ones we adopt by intuition for all tasks in this work) or employing information-theoretic approaches. Further discussions regarding this point can be found in appendix \ref{sec:auto_stage}.

Secondly, the reliance on stage indicators adds a level of inconvenience when directly training RL agents in the real world. 
While it is infrequent to directly train RL agents in the real world due to cost and safety issues, when necessary, stage information can still be obtained using existing techniques, similar to \citep{qt_opt, mt_opt}. 
For example, the “object is grasped” indicator can be acquired by tactile sensors \citep{lin2022tactile, melnik2021using}, and the “object is placed” indicator can be obtained by forward kinematics, visual detection/tracking techniques \citep{qt_opt, mt_opt}, or even large vision-language models \citep{du2023vision}.

\bibliography{iclr2024_conference}
\bibliographystyle{iclr2024_conference}

\newpage

\newpage
\appendix
\onecolumn

\section{Task Descriptions}
\label{sec:task_desc}

For all tasks, we use consistent setups for state spaces, action spaces, and demonstrations. The state spaces adhere to a standardized template that includes proprioceptive robot state information, such as joint angles and velocities of the robot arm, and, if applicable, the mobile base. Additionally, task-specific goal information is included within the state. Please refer to the ManiSkill paper~\citep{gu2023maniskill2} for more details. Below, we present the key details pertaining to the tasks used in this paper.

\subsection{Pick-and-Place}
\begin{itemize}
    \item Stage Indicators:
        \begin{itemize}
            \item Object is grasped: Both of the robot fingers contact the object, and the impulse (force) at the contact points is non-zero.
            \item Object is placed: The distance between the object and the goal position is less than 2.5 cm. This is given by the success signal of the original task, not designed by us.
            \item Robot and object are stationary: The joint velocities of all robot joints are less than 0.2 rad/s. The object velocity is less than 3 cm/s. \textit{This is given by the success signal of the original task, not designed by us.}
        \end{itemize}
    \item Object Set: The objects in training tasks are from the YCB dataset, including 74 objects. And the objects in test tasks are from the EGAD dataset, including around 1600 objects.
    \item Action Space: Delta position of the end-effector and the joint positions of the gripper. 
    \item Demonstrations: We use 100 demonstration trajectories in total for this task family (around 1.4 trajectories per task). The demonstrations are from a trained RL agent.

\end{itemize}

\subsection{Turn Faucet}
\begin{itemize}
    \item Stage Indicators:
        \begin{itemize}
            \item Handle is moving: The joint velocity of the target joint is greater than 0.01 rad/s.
            \item Handle reached the target angle: The joint angle is greater than 90\% of the limit. \textit{This is given by the success signal of the original task, not designed by us.}
        \end{itemize}
    \item Object Set: The objects in training and test tasks are both from the PartNet-Mobility dataset. The training tasks include 10 faucets, and the test tasks include 50 faucets.
    \item Action Space: Delta pose of the end-effector and joint positions of the gripper. 
    \item Demonstrations: We use 100 demonstration trajectories in total for this task family (around 10 trajectories per task). The demonstrations are from a trained RL agent.
\end{itemize}

\subsection{Open Cabinet Door}
\begin{itemize}
    \item Stage Indicators:
        \begin{itemize}
            \item Handle is grasped: Both of the robot fingers contact the handle, and the impulse (force) at the contact points is non-zero.
            \item Door is open enough: The joint angle is greater than 90\% of the limit. This is given by the success signal of the original task, not designed by us.
            \item Door is stationary: The velocity of the door is less than 0.1 m/s, and the angular velocity is less than 1 rad/s. \textit{This is given by the success signal of the original task, not designed by us.}
        \end{itemize}
    \item Object Set: The objects in training and test tasks are both from the PartNet-Mobility dataset. The training tasks include 4 cabinet doors, and the test tasks include 6 cabinet doors. We remove all single-door cabinets in this task family, as they can be solved by kicking the side of the door and this behavior can be readily learned by sparse rewards.
    \item Action Space: Joint velocities of the robot arm joints and mobile robot base, and joint positions of the gripper.
    \item Demonstrations: We use 200 demonstration trajectories in total for this task family (around 50 trajectories per task). The demonstrations are from a trained RL agent.
\end{itemize}

\section{Comparison of Human Effort: Stage Indicators vs. Human-Engineered Rewards }
\label{sec:human_effort_comparison}

This section explains why designing stage indicators is much easier than designing a full dense reward.

The key challenges in reward engineering lies in \textbf{designing reward candidate terms} and \textbf{tuning associated hyperparameters}. To illustrate, let us use the “Open Cabinet Door” task familly as an example. The code of human engineered reward is in Listing \ref{cabinet_human_reward}, and the code of our stage indicators is in Listing \ref{cabinet_stage_indicators}.

The human-engineered reward involves the following reward candidate terms:
\begin{itemize}
    \item Distance between the robot gripper quaternion, and a set of manually designed grasp quaternions
    \item Distance between robot hand and door handle
    \item Signed-distance between tool center point (center of two fingertips) and door handle
    \item Robot joint velocity
    \item Door handle velocity
    \item Door handle angular velocity
    \item Door joint velocity
    \item Door joint position
    \item Multiple boolean functions to determine task stages
\end{itemize}

Each reward candidate term needs 1$\sim$4 hyperparameters (e.g., normalization function, clip upper bound, clip lower bound, scaling coefficient). In total, this reward function involves \textbf{more than 20 hyperparameters} to tune. \textit{The major effort of reward engineering is thus spent iterating over these candidate terms and tuning the hyperparameters by trail and error.} This process is laborious but critical for the success of human-engineered rewards. According to the authors of ManiSkill, they spend \textbf{over one month} crafting the dense reward for the “Open Cabinet Door” tasks.

In contrast, our stage indicators for “Open Cabinet Door” tasks only requires to design \textbf{two boolean functions}: whether the robot has grasped the handle and whether the door is open enough. 
The third stage indicator is given by the tasks success signal so we do not need to design it.
This trims the number of hyperparameters down from 20+ to just  1 (the first boolean function requires one hyperparameter, and the second boolean function is directly taken from the task’s success condition so no hyperparamters), and reduces the lines of code from 100+ to 7 (with a utility function to check grasping, which is from the original codebase).

Therefore, our approach significantly reduces the human effort required for reward engineering.

\begin{lstlisting}[label=cabinet_human_reward, language=Python, caption=Human-engineered rewards for Open Cabinet Door tasks. The code is from the ManiSkill2 github repo (commit id: \texttt{493be36}). ]
    def _compute_grasp_poses(self, mesh: trimesh.Trimesh, pose: sapien.Pose):
        # NOTE(jigu): only for axis-aligned horizontal and vertical cases
        mesh2: trimesh.Trimesh = mesh.copy()
        # Assume the cabinet is axis-aligned canonically
        mesh2.apply_transform(pose.to_transformation_matrix())

        extents = mesh2.extents
        if extents[1] > extents[2]:  # horizontal handle
            closing = np.array([0, 0, 1])
        else:  # vertical handle
            closing = np.array([0, 1, 0])

        # Only rotation of grasp poses are used. Thus, center is dummy.
        approaching = [1, 0, 0]
        grasp_poses = [
            self.agent.build_grasp_pose(approaching, closing, [0, 0, 0]),
            self.agent.build_grasp_pose(approaching, -closing, [0, 0, 0]),
        ]

        pose_inv = pose.inv()
        grasp_poses = [pose_inv * x for x in grasp_poses]

        return grasp_poses

    def _compute_handles_grasp_poses(self):
        self.target_handles_grasp_poses = []
        for i in range(len(self.target_handles)):
            link = self.target_links[i]
            mesh = self.target_handles_mesh[i]
            grasp_poses = self._compute_grasp_poses(mesh, link.pose)
            self.target_handles_grasp_poses.append(grasp_poses)

    def compute_dense_reward(self, *args, info: dict, **kwargs):
        reward = 0.0

        # ----------------------------------------------------- #
        # The end-effector should be close to the target pose
        # ----------------------------------------------------- #
        handle_pose = self.target_link.pose
        ee_pose = self.agent.hand.pose

        # Position
        ee_coords = self.agent.get_ee_coords_sample()  # [2, 10, 3]
        handle_pcd = transform_points(
            handle_pose.to_transformation_matrix(), self.target_handle_pcd
        )
        # trimesh.PointCloud(handle_pcd).show()
        disp_ee_to_handle = sdist.cdist(ee_coords.reshape(-1, 3), handle_pcd)
        dist_ee_to_handle = disp_ee_to_handle.reshape(2, -1).min(-1)  # [2]
        reward_ee_to_handle = -dist_ee_to_handle.mean() * 2
        reward += reward_ee_to_handle

        # Encourage grasping the handle
        ee_center_at_world = ee_coords.mean(0)  # [10, 3]
        ee_center_at_handle = transform_points(
            handle_pose.inv().to_transformation_matrix(), ee_center_at_world
        )
        # self.ee_center_at_handle = ee_center_at_handle
        dist_ee_center_to_handle = self.target_handle_sdf.signed_distance(
            ee_center_at_handle
        )
        # print("SDF", dist_ee_center_to_handle)
        dist_ee_center_to_handle = dist_ee_center_to_handle.max()
        reward_ee_center_to_handle = (
            clip_and_normalize(dist_ee_center_to_handle, -0.01, 4e-3) - 1
        )
        reward += reward_ee_center_to_handle

        # pointer = trimesh.creation.icosphere(radius=0.02, color=(1, 0, 0))
        # trimesh.Scene([self.target_handle_mesh, trimesh.PointCloud(ee_center_at_handle)]).show()

        # Rotation
        target_grasp_poses = self.target_handles_grasp_poses[self.target_link_idx]
        target_grasp_poses = [handle_pose * x for x in target_grasp_poses]
        angles_ee_to_grasp_poses = [
            angle_distance(ee_pose, x) for x in target_grasp_poses
        ]
        ee_rot_reward = -min(angles_ee_to_grasp_poses) / np.pi * 3
        reward += ee_rot_reward

        # ------------------------------------------------- #
        # Stage reward
        # ------------------------------------------------- #
        coeff_qvel = 1.5  # joint velocity
        coeff_qpos = 0.5  # joint position distance
        stage_reward = -5 - (coeff_qvel + coeff_qpos)
        # Legacy version also abstract coeff_qvel + coeff_qpos.

        link_qpos = info["link_qpos"]
        link_qvel = self.link_qvel
        link_vel_norm = info["link_vel_norm"]
        link_ang_vel_norm = info["link_ang_vel_norm"]

        ee_close_to_handle = (
            dist_ee_to_handle.max() <= 0.01 and dist_ee_center_to_handle > 0
        )
        if ee_close_to_handle:
            stage_reward += 0.5

            # Distance between current and target joint positions
            # TODO(jigu): the lower bound 0 is problematic? should we use lower bound of joint limits?
            reward_qpos = (
                clip_and_normalize(link_qpos, 0, self.target_qpos) * coeff_qpos
            )
            reward += reward_qpos

            if not info["open_enough"]:
                # Encourage positive joint velocity to increase joint position
                reward_qvel = clip_and_normalize(link_qvel, -0.1, 0.5) * coeff_qvel
                reward += reward_qvel
            else:
                # Add coeff_qvel for smooth transition of stagess
                stage_reward += 2 + coeff_qvel
                reward_static = -(link_vel_norm + link_ang_vel_norm * 0.5)
                reward += reward_static

                # Legacy version uses static from info, which is incompatible with MPC.
                # if info["cabinet_static"]:
                if link_vel_norm <= 0.1 and link_ang_vel_norm <= 1:
                    stage_reward += 1

        # Update info
        info.update(ee_close_to_handle=ee_close_to_handle, stage_reward=stage_reward)

        reward += stage_reward
        return reward

\end{lstlisting}

\begin{lstlisting}[label=cabinet_stage_indicators, language=Python, caption=Our stage indiactors for Open Cabinet Door tasks\, which is way more easier to design than the human-engineered rewards. ]
def compute_stage_indicators(self):
    stage_indicators = [
        self.agent.check_grasp(self.target_link), # this utility function is given by the original ManiSkill2 codebase. It requires one hyperparameter `max_angle` but we just use the default value
        self.link_qpos >= self.target_qpos, # door is open enough
        # the 3rd stage indicator is just the task success signal, so we don't need to include it here
    ]
    for i in range(1, len(stage_indicators)):
        stage_indicators[i-1] |= stage_indicators[i]
    return stage_indicators

\end{lstlisting}

\section{Comparison with Text2Reward}
\label{sec:compare_text2reward}

Text2Reward \citep{xie2023text2reward} is a concurrent work with our paper. We offer a comparison in this section to help readers understand the differences between our paper and Text2Reward.

While both \citep{xie2023text2reward} and our paper share the common goal of generating rewards for new tasks, they employ fundamentally distinct setups and methodologies. In short, the primary distinction lies in the fact that \textbf{our approach learns rewards from training tasks and success signals (or stage indicators), while \citep{xie2023text2reward} generates rewards based on exemplar reward codes and the knowledge embedded in Large Language Models (LLMs)}.

To elaborate, the following disparities exist in respective setups and assumptions:

\begin{itemize}
    \item Both \citep{xie2023text2reward} and our methods need to interact with environments. However, we emphasize more on evaluating the learned rewards on unseen test tasks.
    \item \citep{xie2023text2reward} assumes access to a pool of instruction-reward code pairs, while our method requires training on relevant training tasks instead.
    \item \citep{xie2023text2reward} assumes access to the source code of the tasks, allowing them to provide LLMs with a Pythonic environment abstraction and various utility functions. In contrast, our method solely relies on success signals (or stage indicators) and does not require the code of the tasks.
\end{itemize}

\section{Implementation Details}
\label{sec:implementation_details}

\subsection{Reward Learning Phase}
\subsubsection{Network Architectures}
\begin{itemize}
    \item Actor Network: 4-layer MLP, hidden units (256, 256, 256)
    \item Critic Networks: 4-layer MLP, hidden units (256, 256, 256)
    \item Discriminator Networks (Reward): 2-layer MLP, hidden units (32)
\end{itemize}

\subsubsection{Hyperparameters}
We use SAC \citep{sac} as the backbone RL algorithm in the reward learning phase of DrS. The related hyperparameters are listed in Table \ref{tab:hyper-reward-learn}.

\subsection{Reward Reuse Phase}
\subsubsection{Network Architectures}
\begin{itemize}
    \item Actor Network: 4-layer MLP, hidden units (256, 256, 256)
    \item Critic Networks: 4-layer MLP, hidden units (256, 256, 256)
\end{itemize}

\subsubsection{Hyperparameters}
During the reward reuse phase, we use different rewards to train agents by SAC \citep{sac}. The related hyperparameters are listed in Table \ref{tab:hyper-reward-reuse}.

\begin{table}[ht]
\parbox{.45\linewidth}{
    \centering
    \begin{tabular}{ll}
    \toprule
    Name                    & Value                           \\
    \midrule
    replay buffer size         & $+\infty$                       \\
    update-to-data (UTD) ratio & 0.5                     \\
    optimizer               & Adam                         \\
    actor learning rate           & 3e-4                           \\
    critic learning rate           & 3e-4                           \\
    discriminator learning rate           & 3e-4                           \\
    target smoothing coefficient & 0.005                  \\
    discount factor & 0.8                     \\
    training frequency & 64 steps                     \\
    target network update frequency & 1 step                  \\
    discriminator update frequency & 1 step                  \\
    batch size & 1024                    \\
    Auto-tune Entropy & True                   \\
    \bottomrule
    \end{tabular}
    \vspace{0.5em}
    \caption{The hyperparameters used in reward learning phase of DrS.}
    \label{tab:hyper-reward-learn}
}
\hfill
\parbox{.45\linewidth}{
    \centering
    \begin{tabular}{ll}
    \toprule
    Name                    & Value                           \\
    \midrule
    replay buffer size         & $+\infty$                        \\
    update-to-data (UTD) ratio & 0.5                     \\
    optimizer               & Adam                         \\
    actor learning rate           & 3e-4                           \\
    critic learning rate           & 3e-4                           \\
    target smoothing coefficient & 0.005                  \\
    discount factor & 0.8                     \\
    training frequency & 64 steps                     \\
    target network update frequency & 1 step                  \\
    batch size & 1024                    \\
    Auto-tune Entropy & True                   \\
    \bottomrule
    \end{tabular}
    \vspace{0.5em}
    \caption{The hyperparameters used in the reward reuse phase of DrS.}
    \label{tab:hyper-reward-reuse}
}
\end{table}

\section{Additional Ablation Study}
\label{sec:additional_ablation}

In this section, we present more ablation studies that are not included in the main paper due to the space limit. These experiments are conducted on the Pick-and-Place task family.

\subsection{Modality of the Inputs to the Rewards}

\label{sec:pcd}

\begin{figure}[ht]
    \centering
    \includegraphics[width=0.45\linewidth]{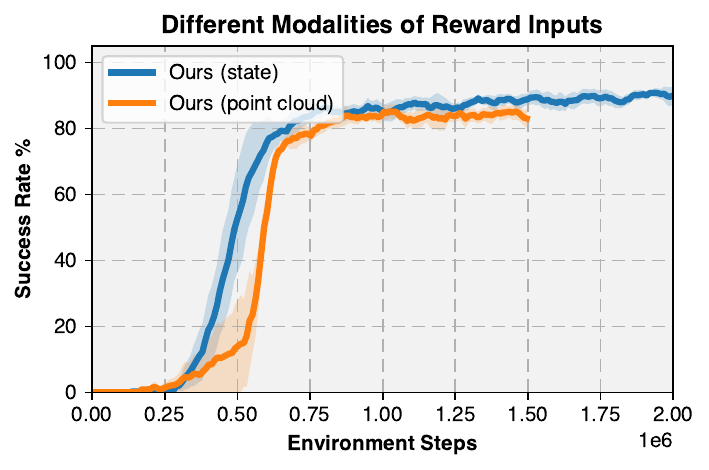}
    \caption{An experiment about using our approach with point cloud inputs, and the point clouds are processed by a PointNet. 
    }
    \label{fig:pcd}

\end{figure}

Our approach is able to accommodate various input modalities for the reward functions, including both low-dimensional state vectors and high-dimensional visual inputs. To demonstrate this compatibility, we conducted an additional experiment using point cloud inputs. In this experiment, the reward function (discriminator) not only considers the low-dimensional state but also takes a point cloud as input, with the point cloud being processed by a PointNet. The results of this experiment are depicted in Fig.~\ref{fig:pcd}.

We can see that the reward with point cloud input performs similarly to the one with state input, which shows that our approach is perfectly compatible with high-dimensional visual inputs. However, the techniques about visual inputs are a bit orthogonal to our focus (reward learning), and learning with visual inputs takes significantly more time, so we still keep most of our experiments on state inputs. 

The results reveal that the reward function utilizing point cloud input performs comparably to the one utilizing state input, demonstrating the seamless integration of our approach with high-dimensional visual inputs. However, it is worth noting that the techniques about visual inputs, while compatible with our framework, are a little bit orthogonal to our focus (reward learning). Moreover, learning with visual inputs typically takes a significantly longer training time. Consequently, the majority of our experiments primarily use state inputs, allowing us to concentrate on the core aspects of reward learning.

\subsection{Discriminator Modification and Stage Indicators}

\label{sec:basic_ablation}

\begin{figure}[ht]
    \centering
    \includegraphics[width=0.7\linewidth]{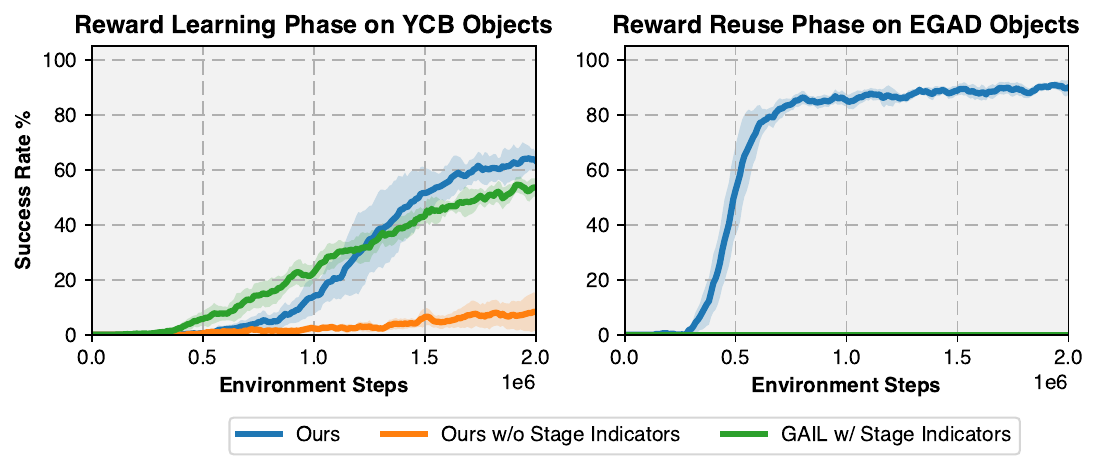}
    \caption{
    An ablation study was conducted to examine the impact of discriminator modification and stage indicators. Both the reward learning phase and reward reuse phase are shown. The learned rewards from the ablated baselines failed to successfully train new agents in the test tasks.
    }
    \label{fig:gail_ablation}
\end{figure}

In contrast to GAIL \citep{gail}, our approach incorporates two critical modifications in the training of discriminators to facilitate the learning of reusable dense rewards. These modifications entail: (a) replacing the agent-demonstration discriminator with the success-failure discriminator, and (b) employing stage indicators by utilizing a separate discriminator for each stage. To ascertain the significance of these modifications, we devised two ablation baselines:

\begin{itemize}
    \item \textbf{GAIL w/ Stage Indicators}: 
    This baseline serves as an equivalent representation of our method without the incorporation of the success-failure discriminator. In GAIL, the discriminator solely distinguishes between agent and expert trajectories, making it incapable of learning separate rewards for each stage. To incorporate the stage indicators within the GAIL framework, we first train the original GAIL on the training tasks. During the reward reuse phase, we linearly combine the GAIL reward with the semi-sparse reward, thus leveraging the stage information. Through experimentation, we explored different weightings to strike an optimal balance between these two reward components.
    \item \textbf{Ours w/o Stage Indicators}: 
    In this baseline, we exclude the stage indicators and solely rely on the task completion signal to train the discriminator. This approach is equivalent to the one-stage reward learning discussed in Sec.~\ref{sec:method_one_stage}.
\end{itemize}

Fig.~\ref{fig:gail_ablation} illustrates the comparison between the two ablation baselines and our method during both the reward learning phase and reward reuse phase. While both "GAIL w/ Stage Indicators" and "Ours w/o Stage Indicators" demonstrate similar success rates as our method at the conclusion of the reward learning phase, it is crucial to emphasize that the learned rewards from both ablation baselines \textbf{fail} to be reused to the test tasks. In contrast, our method achieves the acquisition of high-quality reward functions capable of effectively training new RL agents in the test tasks. This outcome substantiates the indispensability of the two proposed components in facilitating the acquisition of reusable dense rewards.

\subsection{Reward Formulation}
\label{sec:reward_formula}

\begin{figure}[ht]
    \centering
    \includegraphics[width=0.45\linewidth]{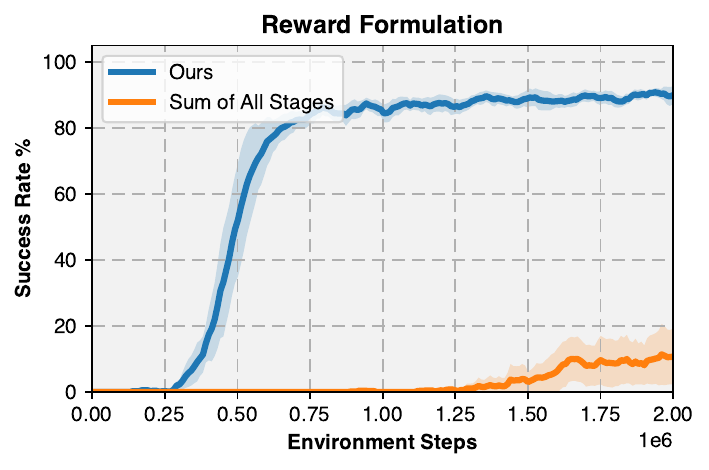}
    \caption{Ablation study of reward formulation. Comparison is done by reusing the learned rewards to train new agents on the test tasks. 
    }
    \label{fig:reward_formulation_ablation}

\end{figure}

In our approach, we leverage the stage indicators and define the reward function as the sum of the semi-sparse reward and the discriminator's bounded prediction for the current stage, as expressed in Eq.~\ref{equ:reward}. This formulation ensures that the reward strictly increases across stages. To evaluate the effectiveness of this formulation, we compare it with a straightforward variant, denoted as $\sum_k \tanh(\text{Discriminator}_k(s'))$, which sums up the discriminator predictions for all stages. As depicted in Fig.~\ref{fig:reward_formulation_ablation}, the simple variant exhibits significantly poorer performance, underscoring the importance of focusing on the dense reward specific to the current stage.

\section{Automatically Generating Stage Indicators}
\label{sec:auto_stage}

This section discusses a few promising solutions to automatically generate stage indicators, drawing inspiration from some recent publications.
Though this topic is \textit{a little bit beyond the scope of our paper}, we believe this is a valuable discussion for the readers.

\subsection{Employ LLMs for Code Generation of Stage Indicators}

Beyond task decomposition, LLMs demonstrate the capability to directly write code \citep{liang2023code, singh2023progprompt, yu2023language, ha2023scaling} for robotic tasks. A recent study \citep{ha2023scaling} exemplified how LLMs, when prompted with the appropriate APIs, can generate success conditions (code snippets) for each subtask. Given the swift advancements in the domain of large models, it is entirely feasible to generate both stage structures and stage indicators using them.

\subsection{Infer Stages via Keyframe Discovery}
The boundaries between stages can be viewed as keyframes in the trajectories. A recent approach introduced by \citep{shi2023waypoint} suggests the automated extraction of such keyframes from trajectories, leveraging reconstruction errors. Given these keyframes, one intuitive solution is to develop a keyframe classifier that can act as a stage indicator. However, this requires a certain degree of consistency across keyframes, and we believe it is an interesting direction to explore.

\newpage

\section{Additional Experiments on Other Domains}

\subsection{Navigation}

\subsubsection{Introduction}

In this section, we incorporated experiments on navigation tasks, which were conducted during the initial stages of our project. We do not include these results in the main paper, as we found these simple navigation tasks to be less interesting compared to the robot manipulation tasks.

\subsubsection{Setup}

\begin{figure}[htbp]
     \centering
     \begin{subfigure}[b]{0.48\textwidth}
         \centering
         \includegraphics[width=\textwidth]{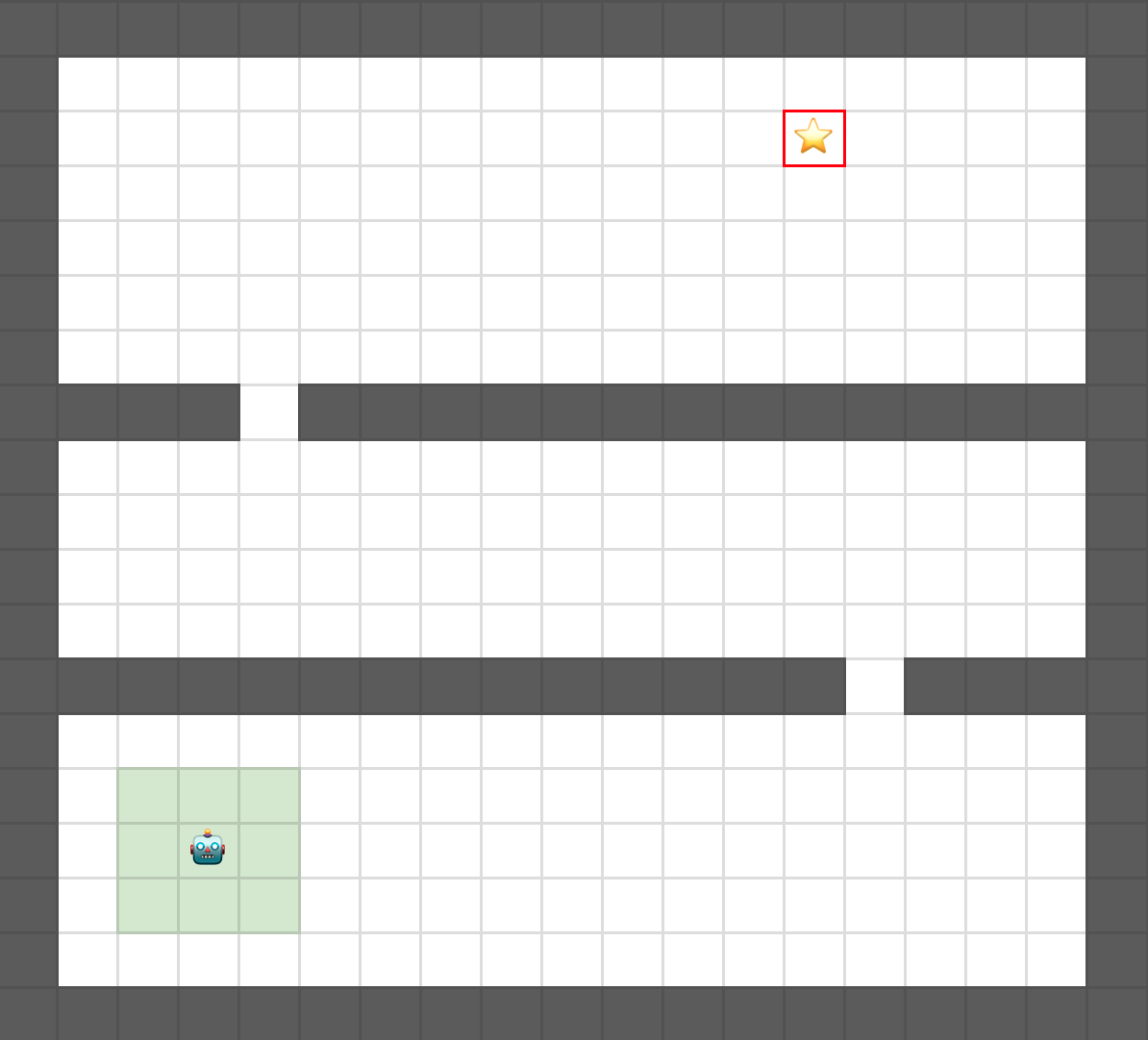}
         \caption{Training Task}
         \label{fig:map_train}
     \end{subfigure}
     \hfill
     \begin{subfigure}[b]{0.48\textwidth}
         \centering
         \includegraphics[width=\textwidth]{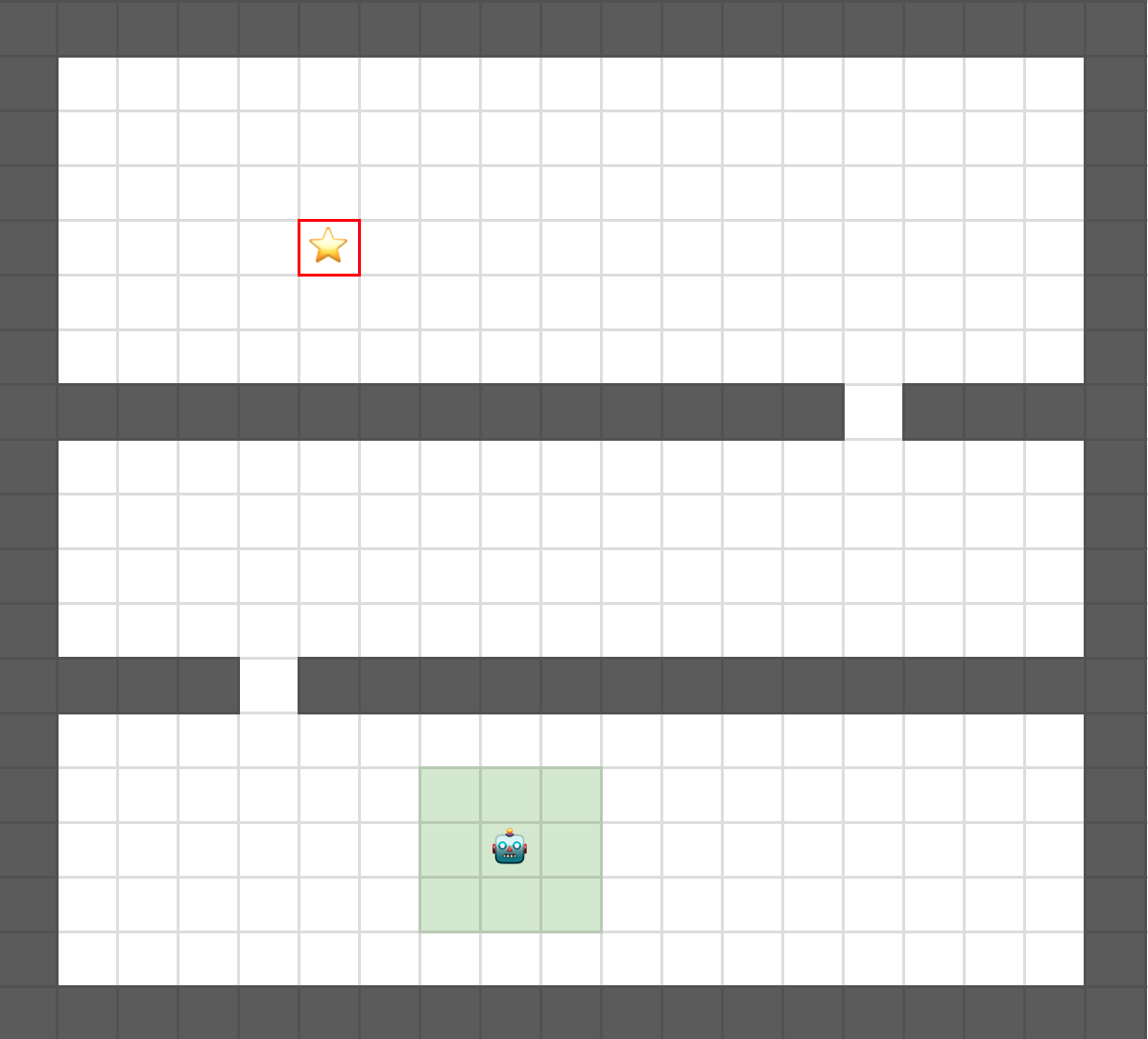}
         \caption{Test Task}
         \label{fig:map_test}
     \end{subfigure}
     \caption{Visualization of the training and test tasks in the navigation domain. The agent begins at a random location in the bottom room, and the goal is randomly positioned in the top room.}
      \label{fig:maps}
      \vspace{-0.5 cm}
\end{figure}

\paragraph{Task Description} We have developed a 2D navigation task conceptually similar to MiniGrid, as visually represented in Fig. \ref{fig:maps}. The maps are 17x17, where the agent is randomly placed in the bottom room and needs to navigate to the star, randomly located in the top room.

\paragraph{Observation} Observations provided to the agent include its xy coordinates, the xy coordinates of the goal, and a 3x3 patch around itself.

\paragraph{Action} The agent has a choice of 5 actions: moving up, down, left, right, or remaining stationary.

\paragraph{Training and Test Set} The reward is learned on the map shown in \ref{fig:map_train} and then reused on the map in \ref{fig:map_test}. The difference between these two maps lies in the positions of two gates.

\subsubsection{Results}
\label{sec:maze_result}

\begin{figure}[htbp]
    \vspace{-1cm}
    \centering
    \includegraphics[width=0.45\linewidth]{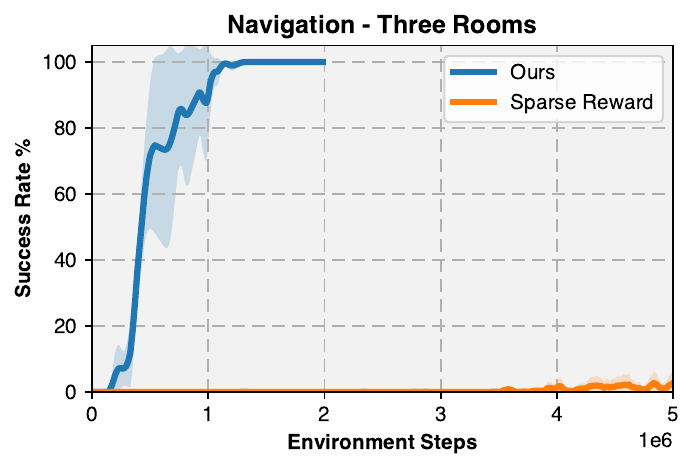}
    \vspace{-0.3 cm}
    \caption{Evaluation results of reusing learned rewards in the navigation task. All curves use DQN to train, but with different rewards. 3 random seeds, the shaded region is std.}
    \label{fig:three_room_result}
\end{figure}

Our method is also effective in learning reusable rewards for navigation tasks. Given the relative simplicity of this specific navigation task, our approach's one-stage version suffices, eliminating the need for additional stage information. The results for this experiment are shown in Fig.~\ref{fig:three_room_result}.

\begin{figure}[htbp]
    \centering
    \includegraphics[width=\linewidth]{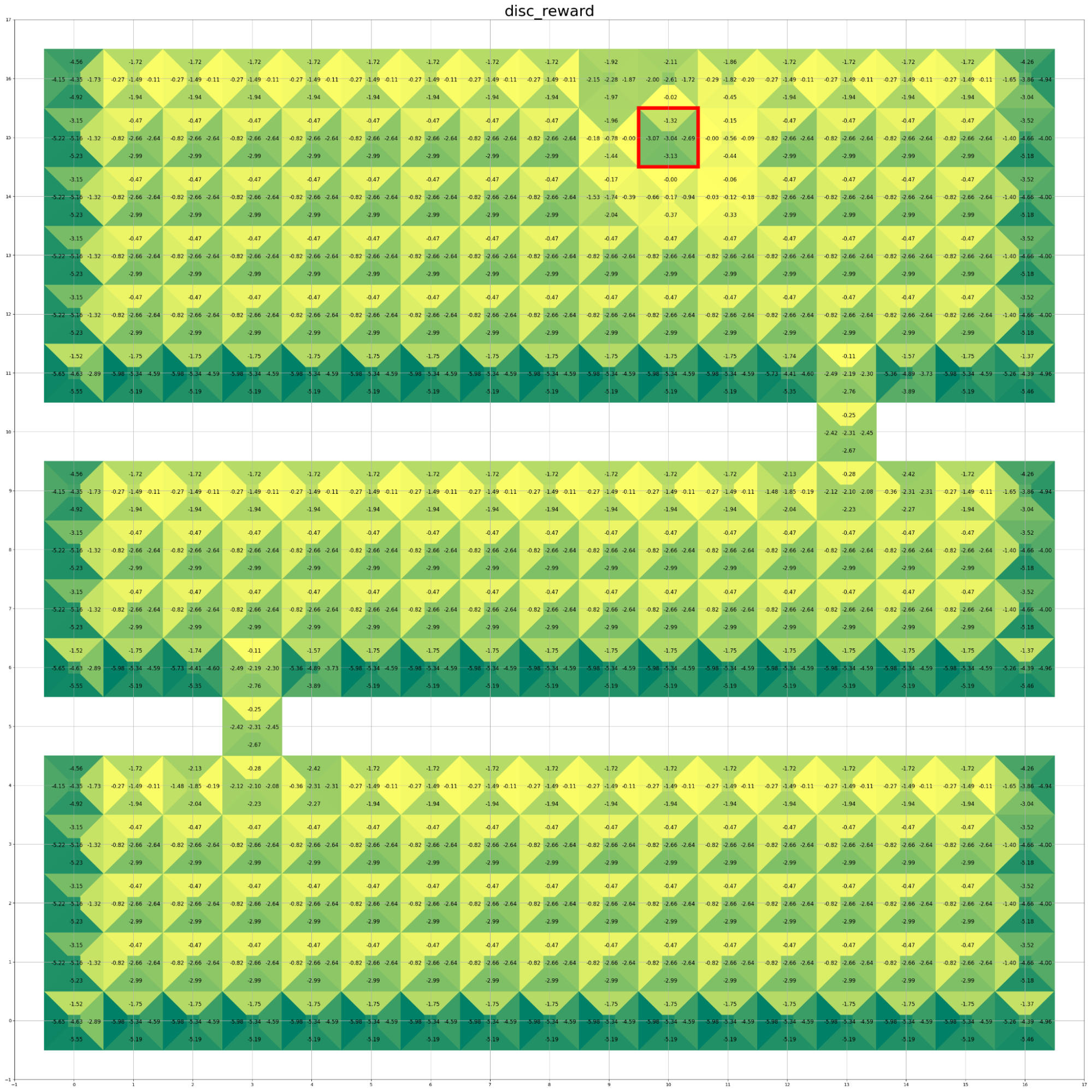}
    \caption{
    Visualization of Learned Reward: Each cell displays five values corresponding to the rewards for five different actions. A lighter color means a higher reward value. The red box shows the location of a randomly chosen goal.
    Note that the inputs to both the learned reward function and the agent include only the local 3x3 area around the agent, excluding any information about the gate positions. Overall, the learned reward encourages upward movement, aligning with the placement of the goal in the top room. When the gates are not within the agent's local 3x3 patch, the rewards for moving left or right are nearly equivalent, which is reasonable since the gate's position cannot be determined without direct observation. However, when the gates are visible within the local patch, the learned reward directs the agent to go through these gates. This behavior of the learned reward aligns well with the task's objectives.
    }
    \label{fig:three_room_reward_vis}
\end{figure}

The results clearly demonstrate that the learned reward from our approach successfully guides the RL agent to complete the task perfectly. In contrast, RL agents with sparse rewards show poor performance. Note that the map used in the test task differs from the training one, so directly transferring policy would not work. We also visualize the learned reward in Fig. \ref{fig:three_room_reward_vis}. See the caption for a detailed analysis.

\newpage

\subsection{Locomotion}

\subsubsection{Introduction}

While it can be tricky to divide locomotion tasks into stages, our method (specifically, the one-stage version) is capable of effectively handling such tasks, if they have a short horizon. In this section, we demonstrate that our approach can learn reusable rewards for Half Cheetah, a representative locomotion task in MuJoCo.

For tasks that are long-horizon and hard to specify stages, such as the \href{https://robotics.farama.org/envs/maze/ant_maze/}{\color{pink}Ant Maze}, crafting rewards is very challenging even for experienced human experts. Therefore, we leave these tasks for future work.

\subsubsection{Setup}

\paragraph{Task Description} Our experiment uses \texttt{HalfCheetah-v3} from Gymnasium. The \texttt{HalfCheetah} task has a predefined reward threshold of 4800, as specified in \href{https://github.com/Farama-Foundation/Gymnasium/blob/main/gymnasium/envs/__init__.py#L277}{\color{pink}their code}, which is used to gauge task completion according to \href{https://gymnasium.farama.org/api/registry/}{\color{pink}their documentation}. Thus, we define the sparse reward (success signal) for this task as achieving an accumulative dense reward greater than 4800.

\paragraph{Training and Test Set} In the reward learning phase, we use the standard \texttt{HalfCheetah-v3} task. In the reward reuse phase, we modify the task by increasing the damping of the front leg joints (thigh, shin, and foot joints) by 1.5 times. This increased damping makes it more challenging for the cheetah to achieve high speeds.

\subsubsection{Results}

\begin{figure}[htbp]
    \centering
    \includegraphics[width=0.45\linewidth]{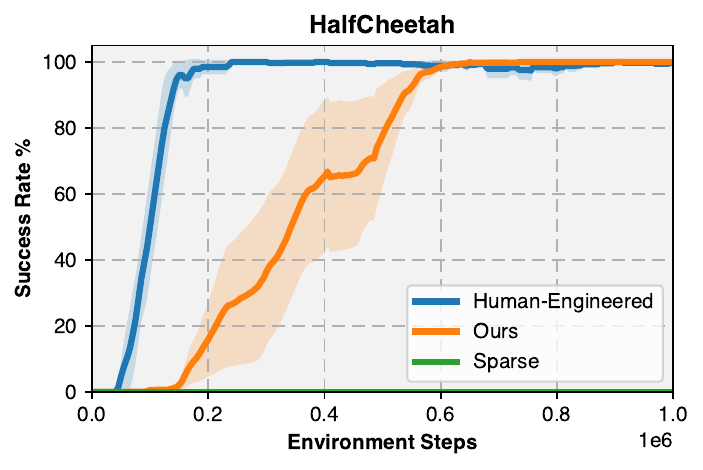}
    \caption{Evaluation results of reusing learned rewards in the \texttt{HalfCheetah-v3} task. All curves use SAC to train, but with different rewards. 3 random seeds, the shaded region is std.}
    \label{fig:halfcheetah_result}
\end{figure}

Our method has successfully demonstrated its capability to learn reusable rewards in the Half Cheetah task. The results are illustrated in Fig.~\ref{fig:halfcheetah_result}. Notably, the performance achieved using the learned reward is comparable to that of the human-engineered reward, while the sparse reward proved ineffective in training an RL agent. Given that many locomotion tasks emphasize low-level control and are typically of a shorter horizon, our approach's one-stage version proves to be highly effective. Additionally, this version does not require any stage information, further underscoring its efficiency and adaptability in handling such tasks.

\newpage

\section{Discussion on the Desired Properties of Dense Rewards}

\subsection{Overview}
Our paper primarily focuses on learning a dense reward, so one important question we want to discuss is: \textit{What kind of dense reward do we aspire to learn?}

It is somewhat challenging to strictly distinguish dense rewards from sparse rewards, due to the lack of strict definitions of dense rewards in the existing literature (to the best of our knowledge). However, this does not preclude a meaningful discussion about the desired properties of dense rewards. Unlike sparse rewards, which typically only provide reward signals when the task is solved, dense rewards offer more frequent and immediate feedback regarding the agent's actions. 

We posit that the fundamental property of an \textit{effective} dense reward is its capacity to enhance the sample efficiency of RL algorithms. The rationale behind this property is straightforward: a well-structured dense reward should \textbf{reduce the need for extensive exploration during RL training}. By providing direct guidance and immediate feedback, the agent can quickly discover optimal actions, thereby accelerating the learning process.

In line with this philosophy, an \textbf{ideal dense reward should allow the derivation of optimal policies with minimal effort}. By analyzing a simple tabular case, we find that \textbf{our learned reward exhibits this great property}. To be more specific, in the example below, \textbf{we can obtain the optimal policy by greedily following the path of maximum reward at each step}.

\subsection{Analysis on a Simple Tabular Case}

Under certain assumptions, we can obtain the optimal policy by greedily following the path of maximum reward at each step, i.e.,
$$
\pi^*(s)=\argmax_a R^{\dagger}(s,a)
$$
, where $\pi^*$ is the optimal policy and $R^{\dagger}$ is the learned reward.

\subsubsection{Setup and Assumptions}
In this analysis, we consider a MDP with the following assumptions:
\begin{itemize}
    \item Deterministic transitions: $s'=P(s,a)$
    \item Discrete and finite state/action space: $S=\{s_0,s_1,...\}$, $A=\{a_0,a_1,...\}$ 
    \item Given sparse reward: $R(s,a,s')=1$ if $s'=s_{goal}$, otherwise 0
    \item Discount factor: $\gamma<1$
\end{itemize}

Other assumptions about our approach:
\begin{itemize}
    \item Only one stage, so the one-stage version of our approach is applied.
    \item The buffers for success trajectories and failure trajectories are large enough, but not infinite.
    \item After training for a sufficiently long time, policy converges to the optimal policy $\pi^*$. (This is a strong assumption, but it is possible in theory.)
\end{itemize}

\subsubsection{Notations}
\begin{itemize}
    \item Learned reward: $R^{\dagger}(s,a)=\tanh(\text{Discriminator}(s,a))$, so $R^{\dagger}(s,a) \in (-1,1)$
    \item Buffer for success trajectories $\mathcal{B}^+$, buffer for failure trajectories $\mathcal{B}^-$
    \item Optimal policy: $\pi^*(a|s)$, which represents the probability of choosing action $a$ at state $s$. Here we overload the notation $\pi^*$ to capture the potential multi-modal output of the policy.
\end{itemize}

\subsubsection{Connection between Optimal Policy and Learned Reward}

Here, we want to demonstrate that \textit{the learned reward of an optimal action is always higher than that of any non-optimal action in each state}. If this holds, it then becomes feasible to straightforwardly identify the optimal action at each state by adopting a greedy strategy that selects the action yielding the highest reward.

When $\gamma<1$, $\pi^*$ will go to $s_{goal}$ by the shortest paths, so $\pi^*(a|s)=1/k_s$ or 0, where $k_s$ is the number of optimal actions at $s$.

$\forall s$, there are two kinds of actions $a^+$ and $a^-$

\begin{enumerate}
    \item $\pi^*(a^+|s)>0$, which means $a^+$ is one of the optimal actions. 
   Then $(s,a^+)$ must be in $\mathcal{B}^+$, possibly be in $\mathcal{B}^-$.
   Therefore, $R^{\dagger}(s,a^+)> -1 + \epsilon$, when the discriminator converges. This is because the buffers are finite-size, $(s,a^+)$ will be sampled into positive training data of the discriminator with a probability larger than 0.
   \item $\pi^*(a^-|s)=0$, which means $a^-$ is NOT one of the optimal actions.
   Then $(s,a^-)$ will only be in $\mathcal{B}^-$, and will NOT be in $\mathcal{B}^+$
   Therefore, $R^{\dagger}(s,a^-)\to -1$, when the discriminator converges. This is because $(s,a^-)$ will only show in the negative training data of the discriminator.
\end{enumerate}

Therefore, we have $R^{\dagger}(s,a^+) > R^{\dagger}(s,a^-)$ for all states $s$. By employing a greedy strategy that selects $\argmax_a R^{\dagger}(s,a)$, we can reach the goal states in the same way as how the optimal policy $\pi^*$ reaches the goal.

\subsection{Further Discussions}

This subsection is dedicated to addressing additional questions the readers may raise after reading the above analysis.

\subsubsection{Does the above conclusion generalize to more complicated cases?}

Although our analysis highlights a desirable property of the learned reward in a simple tabular case, this finding should not be hastily generalized to more complex cases, such as the robotic manipulation tasks used in our paper. This caution is due to two primary reasons:

\begin{enumerate}
    \item In environments where the state and action spaces are continuous, the ability of the neural network to interpolate plays a significant role in shaping the final learned reward.
    \item Practically, achieving convergence for both the policy and the discriminator can be a very time-consuming process.
\end{enumerate}

\subsubsection{The Necessity of Learned Reward Despite Its Similarity to Policy}

The learned reward might appear redundant at first glance, as it seems to convey the same information as the learned policy. This observation raises a potential question: why is there a need for a learned reward if we already have a learned policy? Couldn't we just utilize the learned policy directly?

The answer lies in the distinct advantages that the learned reward offers, particularly when adapting to new tasks. When the environment dynamics change, a new policy can be effectively retrained using the learned reward in conjunction with the new environmental dynamics. Directly transferring the policy, or fine-tuning it with a sparse reward, can be less efficient in certain situations. 
For a practical illustration of this concept, refer to Fig. \ref{fig:three_room_reward_vis} and Sec, \ref{sec:maze_result}. These sections provide a compelling example where the transfer of rewards demonstrates success, in contrast to the less effective transfer of policies.

\end{document}